\documentclass{article}

\usepackage{microtype}
\usepackage{graphicx}
\usepackage{subcaption}
\usepackage{multirow}
\usepackage{booktabs} 
\usepackage{makecell}
\usepackage{arydshln}
\usepackage{pifont}

\usepackage{hyperref}

\usepackage[preprint]{icml2026}

\usepackage{amsmath}
\usepackage{amssymb}
\usepackage{mathtools}
\usepackage{amsthm}

\usepackage[capitalize,noabbrev]{cleveref}

\theoremstyle{plain}

\theoremstyle{definition}

\theoremstyle{remark}

\usepackage{makecell}

\usepackage[textsize=tiny]{todonotes}

\begin{document}

\twocolumn[
  \icmltitle{EAQuant: Enhancing Post-Training Quantization for MoE Models via Expert-Aware Optimization}



  \icmlsetsymbol{equal}{*}

  \begin{icmlauthorlist}
    \icmlauthor{Zhongqian Fu}{equal,yyy}
    \icmlauthor{Tianyi Zhao}{equal,yyy,sch}
    \icmlauthor{Ning Ding}{yyy}
    \icmlauthor{Xianzhi Yu}{yyy}
    \icmlauthor{Xiaosong Li}{yyy}
    \icmlauthor{Yehui Tang}{yyy}
    \icmlauthor{Yunhe Wang}{yyy}
    \textrm{\{fuzhongqian\}@huawei.com, \{ty\_zhao\}@buaa.edu.cn}
  \end{icmlauthorlist}

  \icmlaffiliation{yyy}{Huawei Noah's Ark Lab}
  \icmlaffiliation{sch}{Beihang University}

  \icmlcorrespondingauthor{Yehui Tang}{yehui.tang@huawei.com}
  \icmlcorrespondingauthor{Yunhe Wang}{yunhe.wang@huawei.com}

  \icmlkeywords{Machine Learning, ICML}

  \vskip 0.3in
]



\printAffiliationsAndNotice{\icmlEqualContribution}

\begin{abstract}
  Mixture-of-Experts (MoE) models enable scalable computation and performance in large-scale deep learning but face quantization challenges due to sparse expert activation and dynamic routing. Existing post-training quantization (PTQ) methods fail to address activation outliers, routing instability, and sparse expert calibration, leading to significant performance degradation. To address this, we propose EAQuant, a PTQ framework tailored for MoE architectures. Our method introduces three expert-aware innovations: (1) \textbf{smoothing aggregation} to suppress activation outliers, (2) \textbf{routing consistency alignment} to preserve expert selection post-quantization, and (3) \textbf{calibration data balance} to optimize sparsely activated experts. These strategies collectively enable robust, high-precision quantization of MoE models under ultra-low-bit constraints.
  Extensive experiments across several extreme quantization settings (e.g., W4A4/W3A4/W3A3/W2A4) demonstrate that EAQuant significantly outperforms existing methods, achieving average accuracy improvements of 1.15 - 13.81\% across three diverse MoE architectures, with particularly pronounced gains in reasoning tasks and robust performance retention under aggressive quantization. By integrating these innovations, EAQuant establishes a new state-of-the-art for high-precision, efficient MoE model compression. 
  Our code is available at \url{https://github.com/darren-fzq1/EAQuant}.
\end{abstract}

\section{Introduction}
Recent advancements in large language models (LLMs) have been driven by the Mixture-of-Experts (MoE) architecture~\cite{muennighoff2025olmoe,jiang2024mixtralexperts,xue2024openmoe,qwen15moe,dai2024deepseekmoe,liu2024deepseekv3}, which employs dynamic routing to scale parameters efficiently with sublinear computational growth. By activating only a subset of expert networks based on input requirements, MoE enables larger models within limited computational budgets. However, deployment challenges persist due to high memory and computational overhead~\cite{Cai_2025,liu2025survey}. The requirement to load all experts into memory simultaneously, even when only a few are active, creates memory bandwidth bottlenecks and increases inference costs. These issues highlight the need for effective compression techniques to reduce resource demands for practical deployment on constrained devices.

\begin{figure*}
	\centering
	\includegraphics[width=\linewidth]{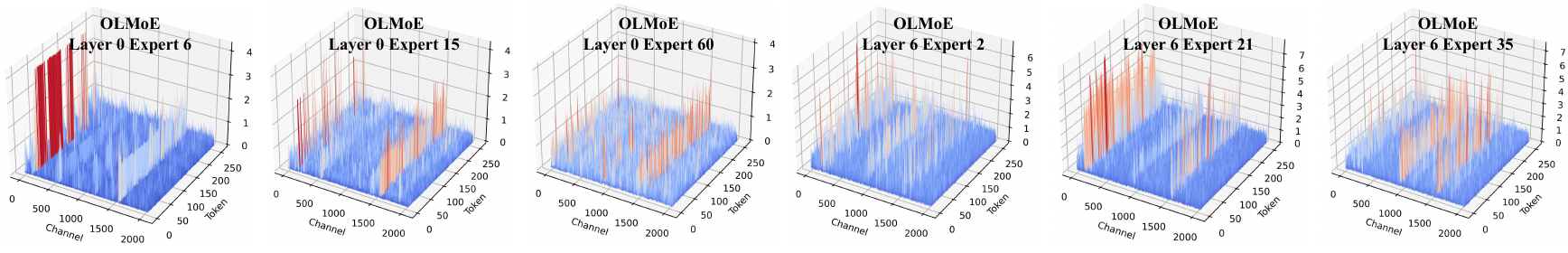}
	\vspace{-2em}
	\caption{Activation outliers across channels in different experts and layers of OLMoE model, exhibiting similar distribution patterns to dense models and posing challenges for quantization. Activation visualizations for other models will be provided in the Appendix.}
	\label{fig_Activation_distribution}
	\vspace{-1em}
\end{figure*} 

\begin{figure*}
	\centering
	\includegraphics[width=\linewidth]{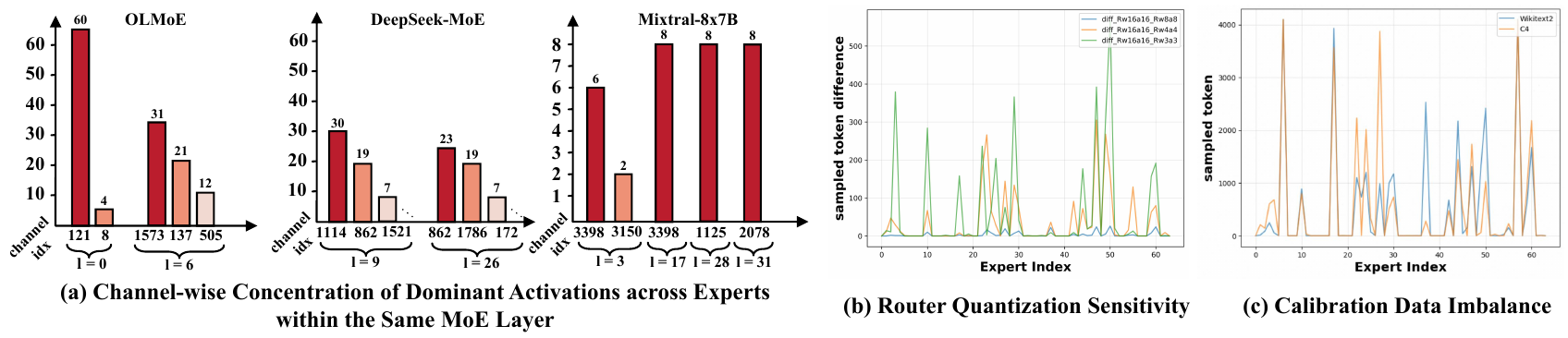}
	\vspace{-2em}
	\caption{Empirical observations motivating expert-aware MoE quantization. More results for other models are provided in the Appendix.}
	\label{fig_observation}
	\vspace{-1.5em}
\end{figure*}

Post-Training Quantization (PTQ) has demonstrated remarkable efficacy in reducing model size and memory footprint. 
\ding{182} Severe activation outliers persist across different experts and layers in MoE architectures, as illustrated in Figure~\ref{fig_Activation_distribution}, 
posing great challenges to quantization.
Specifically, applying channel-wise smoothing~\cite{xiao2023smoothquant,wei2023osplus,shao2023omniquant} independently to each expert generates scaling factors that cannot be absorbed by preceding normalization layers, inevitably introducing additional computational overhead.
Conversely, coarse-grained solutions that concatenate experts into global entities, though compatible with dense model implementations, rely on global maximum statistics that overlook the varying quantization difficulties across experts, ultimately limiting performance gains. 
\ding{183} More critically, unlike dense models, the routing layers unique to MoE architectures exhibit extreme sensitivity to quantization perturbations, as illustrated in Figure ~\ref{fig_observation}(b), where minute deviations in gating scores can distort the top-$k$ selection logic, causing token misrouting and cascading performance degradation. 
\ding{184} Additionally, as depicted in Figure~\ref{fig_observation}(c), the inherent sparsity of expert activations leads to uneven calibration data coverage, resulting in inaccurate quantization parameter estimation for low-frequency experts. 
While MoEQuant~\cite{hu2025moequant} attempts to alleviate this issue through expert-balanced self-sampling, such generative calibration approaches may compromise fair comparisons with other baselines. 
Existing PTQ methods~\cite{frantar2023gptq,lin2023awq,huang2025MCMOE,li2025quantmoebench,zheng2025moqa} generally treat routing layers as passive components, neglecting their intricate interplay with expert activations while overlooking the activation sparsity that underlies the MoE architecture's efficiency advantages. Consequently, these critical challenges remain inadequately addressed.

To tackle these challenges, we propose a novel \textbf{E}xpert-\textbf{A}ware Post-Training \textbf{Quant}ization (EAQuant) method. 
Specifically, we introduce an \textbf{expert-aware smoothing aggregation} strategy to suppress activation outliers across MoE experts. This strategy first computes fine-grained smoothing coefficients for individual experts, then aggregates them into a unified channel-wise smoothing vector. By aggregating scaling requirements from both experts and router, our approach redistributes outlier magnitudes while maintaining mathematical equivalence through parameter fusion with preceding normalization layers. This design preserves expert-specific quantization characteristics while enabling adaptive distribution smoothing across experts. 
To ensure consistent expert selection post-quantization, we introduce \textbf{expert-aware routing consistency alignment} through a dual-objective calibration process. This minimizes both logit reconstruction error and Kullback-Leibler divergence between full-precision and quantized routing probabilities, guaranteeing stable top-k expert activation despite quantization-induced perturbations.  
Finally, we resolve expert-level activation sparsity through \textbf{expert-aware calibration data balance}, where underutilized experts receive prioritized sampling from augmented datasets until their activation counts meet parity with computationally derived expectations. This addresses imbalanced expert utilization and ensures robust performance across all experts.  
By integrating these three components, our method achieves significant quantization error reduction without introducing additional computational overhead, offering a novel solution for efficient MoE model deployment.

Extensive evaluations across diverse MoE architectures and quantization configurations demonstrate that EAQuant achieves state-of-the-art~(SOTA) performance under ultra-low-bit constraints. For instance, EAQuant improves average task accuracy by 1.37\%, 1.15\%, and 1.15\% over the SOTA method DuQuant across OLMoE-7B, DeepSeek-MoE-16B, and Mixtral-8x7B under W4A4 quantization, with particularly pronounced gains in reasoning benchmarks (e.g., +2.52\% on ARC-E for Mixtral-8x7B) and closer perplexity alignment to full-precision baselines. 
Critically, EAQuant exhibits superior robustness in aggressive quantization regimes. Under W3A4, it achieves up to 2.28\% average accuracy improvement and under extreme settings like W3A3 and W2A4, where DuQuant suffers severe degradation, EAQuant maintains stable perplexity while delivering substantial accuracy gains (e.g., +13.81\% average accuracy on Mixtral-8x7B). These advancements stem from our expert-aware smoothing aggregation strategy,  expert-aware routing consistency alignment, and expert-aware calibration data balance, collectively establishing EAQuant as new benchmark for efficient, high-precision MoE quantization.

\section{Method}
\begin{figure*}
	\centering
	\includegraphics[width=0.95\linewidth]{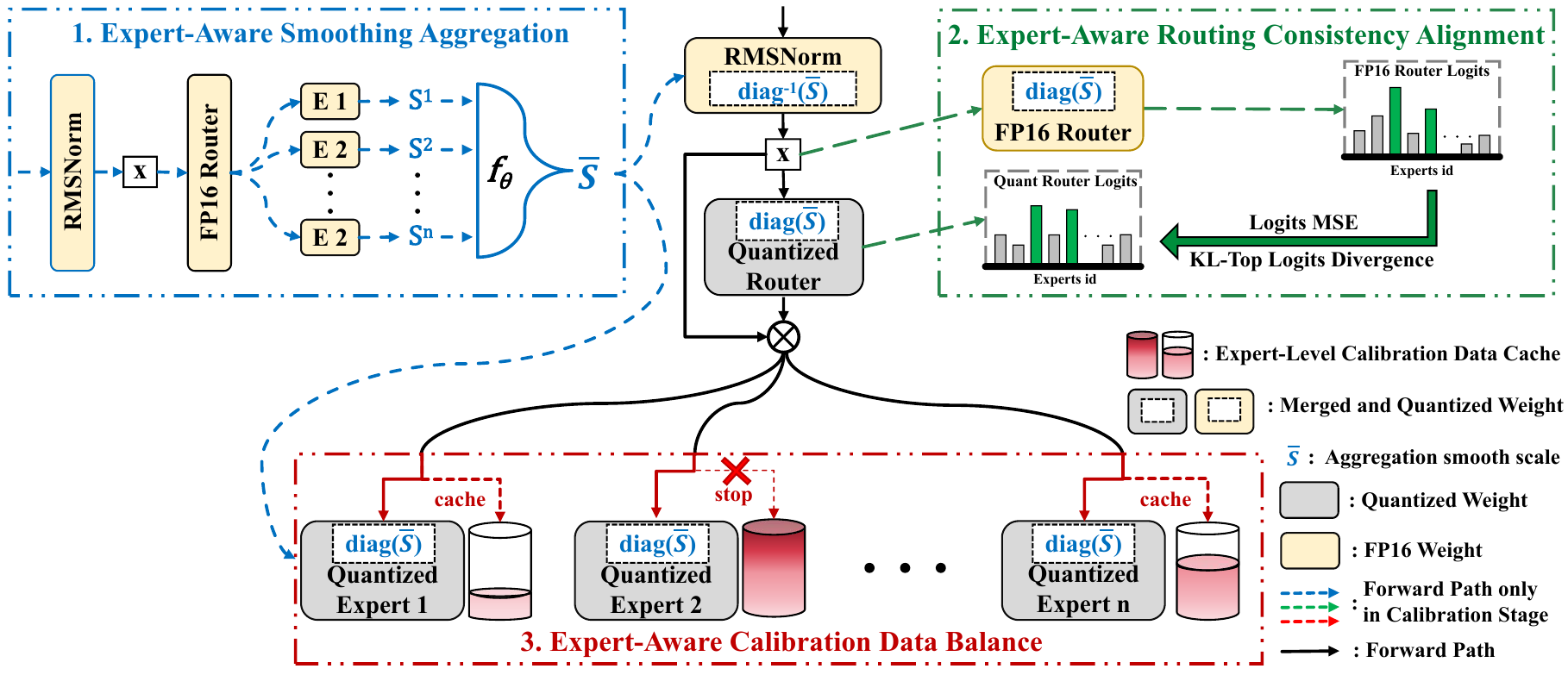}
	\caption{The overview of our proposed EAQuant method with three key components. 1) Expert-Aware Smoothing Aggregation. 2) Expert-Aware Routing Consistency Alignment. 3) Expert-Aware Calibration Data Balance.}
	\label{fig2_EAQuant}
	\vspace{-1em}
\end{figure*}

In this section, we present EAQuant, a post-training quantization (PTQ) method for Mixture-of-Experts (MoE) architectures. As shown in Figure~\ref{fig2_EAQuant}, our approach addresses three core challenges in MoE quantization: activation outlier patterns, quantization sensitivity of MoE router, and calibration data imbalance for rare experts. To resolve these issues, we propose three expert-aware strategies: (1) smoothing aggregation to suppress activation outliers and stabilize expert participation, (2) routing consistency alignment to align pre- and post-quantization router logit distributions and maintain selection consistency, and (3) calibration data balancing to address uneven activation frequencies of local experts during calibration. These strategies directly target the challenges while preserving MoE system characteristics.

\subsection{Motivation}
\ding{182}~\textbf{Activation Outlier Patterns in MoE.}
Modern Mixture-of-Experts (MoE) models exhibit persistent activation outliers across experts and layers. As shown in Figure~\ref{fig_Activation_distribution}, maximum activation values concentrate in a small subset of channels across experts within the same layer, forming a sparse yet statistically homogeneous pattern. This is further validated in Figure~\ref{fig_observation}(a), where activation outliers demonstrate cross-expert consistency despite diverse expert parameterizations. These spatially concentrated and inter-correlated outliers suggest structural properties of MoE dynamics rather than random artifacts. This observation motivates expert-aware aggregation strategies that capture global outlier patterns by leveraging shared distribution characteristics, avoiding information loss from uniform aggregation. Our method explicitly models the interplay between outlier concentration and expert specialization to enhance MoE efficiency.

\ding{183}~\textbf{Quantization Sensitivity of MoE Router.}
The routing layer in Mixture-of-Experts (MoE) architectures dynamically assigns tokens to experts but is highly sensitive to quantization granularity. Quantization perturbs routing logits, causing two key issues: structural instability (unpredictable shifts in expert utilization) and dynamic workload imbalance (uneven token distribution). As shown in Figure~\ref{fig_observation}(b), lower bitwidths amplify token distribution disparities, with some experts receiving disproportionately fewer tokens while others become overburdened. These imbalances degrade accuracy and undermine MoE load-balancing efficiency. Existing MoE quantization methods focus on weight/activation compression~\cite{hu2025moequant,huang2025MCMOE}, neglecting the routing layer’s vulnerability. To address this, we propose expert-aware routing consistency alignment via a dual-objective calibration process that minimizes logit reconstruction error and Kullback-Leibler divergence between full-precision and quantized routing probabilities. This ensures stable top-k expert activation under quantization, preserving accuracy and load-balancing efficiency in low-precision MoE implementations.

\ding{184}~\textbf{Calibration Data Imbalance for Rare Experts.}
MoE models exhibit power-law activation distributions, where a small subset of “core” experts handle majority of tokens, leaving “niche” experts underutilized. During PTQ calibration, rare experts receive insufficient data coverage, causing their quantization parameters (e.g., scaling factors) to overfit to outliers or noise.
As shown in Figure~\ref{fig_observation}(c), we plot the sample distribution on the first MoE layer of OLMoE-7B, this imbalance manifests consistently across both calibration sets.
Current methods~\cite{kim2023moqe,huang2025MCMOE,li2025quantmoebench} almost ignore this sparsity, which compromises the MoE’s adaptive computation advantage. 
MoEQuant~\cite{hu2025moequant} proposes expert-balanced self-sampling to create a balanced calibration dataset, however generating new calibration data may compromise the fairness of comparison with other methods.
Therefore, this calibration sparsity remains unaddressed in sota PTQ methods, creating a critical barrier to efficient MoE deployment.

\subsection{Expert-Aware Smoothing Aggregation}
\label{subsec:Smoothing}
Existing literature regarding the quantization for Mixture-of-Experts models mainly focus on quantizing weights only, while the activations still remain floating-point values. Our method efficiently quantizing both activations and weights by solving two major challenges encountered when quantizing activations of MoE model.

In post-training quantization for large language models, a small number of channels in activation tensors usually exhibit abnormal values with extremely large magnitude. Some well-known works like SmoothQuant~\cite{xiao2023smoothquant} and OmniQuant~\cite{shao2023omniquant} utilize the technique of mergeable smoothing vector to scale the dynamic range of activation tensor before quantizing activations during inference. Specifically, the smoothing vector $s$ is computed per-channel by $s_j = \max(|\mathbf{x}_j|)^\alpha/ \max(|\mathbf{W}_j|)^{1-\alpha}$ to alleviate the outlier value by channel-wise scaling $\mathbf{\tilde x}=\mathbf{x}\cdot\text{diag}^{-1}(s)$, and therefore mitigate the quantization difficulty. Moreover, the smoothing vector $s$ can be merged into the preceding normalization layer, incurring no extra computation overhead.

However, this technique faces a critical generalizability issue when quantizing activations of MoE model.
For a token vector $\mathbf{x}\in \mathbf{R}^d$ with $d$ channels, and an MoE layer with $n$ local experts. The final output of the layer is the weighted sum of selected experts' local outputs by the gate values:
{
\begin{equation}
   \mathbf  y = \sum_{i\in\mathcal{T}} p^i(\mathbf x)E^i(x),
\end{equation}
}
where $\mathcal{T}$ is the set of indices with the highest top-k gate values.
In this situation, the original activation smoothing method requires per-expert smoothing vectors $\{s^i\in\mathbf{R}^d\}_{i=1}^n$ before quantizing activations, respectively computed as:
{
\begin{equation}
s^i_j = \frac{\max(|\mathbf{x}^i_j|)^\alpha}{\max(|\mathbf{W}^i_j|)^{1-\alpha}} \quad \forall j \in \{1,2,\cdots,d\},
\end{equation}
}
where subscript $j$ denotes the $j$-th input channel, $\mathbf{W}^i$ represents the first weight matrix of the $i$-th local expert, and $\mathbf{x}^i$ corresponds to the input activation vector routed to the $i$-th local expert by the router mechanism.

\newcommand{\ie}{\textit{i.e.}}

While the weight transformation $\mathbf{\tilde W}^i = \mathbf{W}^i  \text{diag}(s^i)$ preserves mathematical equivalence through $\mathbf{x}^i\mathbf{W}^i = (\mathbf{x}^i~ \text{diag}^{-1}(s^i))\cdot(\text{diag}(s^i)\mathbf{W}^i)$, the activation scaling operation $\mathbf{x}^i~\text{diag}^{-1}(s^i)$ must be dynamically executed \textit{after} expert selection, introducing $\mathbf{O}(kd)$ computational overhead per token where $k$ is the number of experts each token is routed to. The reason is that the preceding normalizing layer (\ie~RMSNorm and LayerNorm) can only absorb one vector before inference.

Our key point is to construct a unified smoothing vector $\overline{s}$ that satisfies
{
\begin{equation}
\overline{s}_j \gtrsim \max_{i \in [1,n]}(s_j^i), \quad \forall j \in \{1,2,\cdots,d\},
\end{equation}
}
to suppress channel-related extreme values in activation regardless of the routing destination. We achieve this through a generalized channel-wise aggregation operator 
~$\mathbb{A}$~(e.g., maximum) over expert-specific requirements:
{
\begin{equation}
\overline{s}_j = \mathbb{A}_{i \in [1,n]} (s_j^i).
\end{equation}
}
This aggregation guarantees that for any selected expert \textit{i}, we have
\begin{equation}
\overline{s}_j \gtrsim s^i_j \Rightarrow \text{diag}^{-1}(\overline{s}) \preceq \text{diag}^{-1}(s^i),
\end{equation}
where $\preceq$ denotes element-wise inequality, ensuring numerical stability when quantizing the activation with outlier channels.
During the forward propagation of the MoE module, the router's weight 
$\mathbf{W}^{\text{gate}}$ operates on the original input activation, while each local expert receives a subset of features routed by the mechanism.
Therefore we extend our unified smoothing vector to incorporate router weights $\mathbf{W}^{\text{gate}} \in \mathbb{R}^{d \times n}$ by introducing router-specific scaling vector 

$s^{\text{gate}}=\frac{\max(|\mathbf{x}_j|)^{\alpha}}{\max(|\mathbf{W}^{\text{gate}}_j|)^{1-\alpha}}$ into the aggregation process:
{
\footnotesize
\begin{equation}
\overline{s}_j = \max\left(\underbrace{\mathbb{A}_{i \in [1,n]} \left(\frac{\max(|\mathbf{x}^i_j|)^{\alpha}}{\max(|\mathbf{W}^i_j|)^{1-\alpha}}\right)}_{\text{Expert requirements}}, ~\underbrace{\frac{\max(|\mathbf{x}_j|)^{\alpha}}{\max(|\mathbf{W}^{\text{gate}}_j|)^{1-\alpha}}}_{\text{Router requirement}} \right)
\end{equation}
}
This joint maximization guarantees $\overline{s}_j\geq \max\left(s^{\text{gate}}_j,\{s^i_j\}_{i=1}^n\right).$
The unified scaling enables equivalent transformations for both expert and router computations:
{\footnotesize
\begin{equation}
\begin{aligned} 
\text{Router~:~} \mathbf{\tilde x}\mathbf{\tilde W}^{\text{gate}} & = (\mathbf{x} \text{diag}^{-1}(\overline{s}))(\text{diag}(\overline{s}) \mathbf{W}^{\text{gate}} )=\mathbf{x}\mathbf{W}^{\text{gate}}, \\
\text{Expert~i~:~}\mathbf{\tilde x}^i\mathbf{\tilde W}^i & = (\mathbf{x}^i \text{diag}^{-1}(\overline{s}))(\text{diag}(\overline{s}) \mathbf{W}^i )=\mathbf{x}^i\mathbf{W}^{i}. 
\end{aligned}
\label{eq:smooth}
\end{equation}}
And we can absorb $\overline{s}$ into the preceding RMSNorm layer through parameter fusion

{
\begin{equation}
 \mathbf{\tilde x}=\text{RMSNorm}'(\mathbf{x}) = \frac{\gamma \oslash \overline{s}}{\sqrt{\frac{1}{d}\sum_{j=1}^d \mathbf{x}_j^2 }} \odot \mathbf{x}.
\end{equation}
}
\subsection{Expert-Aware Routing Consistency Alignment}
\label{subsec:router_align}

Accurate expert routing is critical to the performance of MoE models. However, router decisions are notoriously fragile under quantization noise, even small perturbations in router logits can alter the top-$k$ expert selection, leading to misrouted tokens and cascading performance degradation.
To address this issue, we propose an expert-aware routing consistency alignment strategy that explicitly preserves routing behavior during post-training quantization. Let $\mathbf{W}^{\text{gate}} \in \mathbb{R}^{d \times n}$ denote the router weight matrix after expert-aware smoothing aggregation (Section~\ref{subsec:Smoothing}), and let $\mathbf{\tilde x}$ be the corresponding normalized input activation.
We adopt a uniform quantization operator
\begin{equation}
\mathcal{Q}(\mathbf{W}) = \mathrm{clip}\!\left( \mathrm{round}\!\left(\frac{\mathbf{W}}{\Delta}\right) + z,\; q_{\min},\; q_{\max} \right),
\end{equation}
where $\Delta$ and $z$ are the scale and zero-point parameters to be calibrated. Our calibration process solves:
{
\footnotesize
\begin{equation}
\min_{\theta } 
\underbrace{\mathbb{E}_{\mathbf{\tilde x}}
\!\left[\| \mathbf{\tilde x}\mathbf{\tilde W}^{\text{gate}} -
\mathbf{\tilde x}\mathcal{Q}(\mathbf{\tilde W}^{\text{gate}}) \|_2^2\right]}_{\text{Logit MSE}}
+
\underbrace{\mathbb{E}_{\mathbf{\tilde x}}
\!\left[D_{\text{KL-Top}}(p_{\text{fp}} \| p_{\text{quant}})\right]}_{\text{Routing KL Divergence}},
\end{equation}
}
where $\theta$ represents quantization parameters such as scale and zero-point for $\mathbf{W}^{\text{gate}}$.
We introduce the KL-Top divergence proposed in~\cite{hu2025ostquant}, which restricts KL regularization to a subset of routing-relevant experts rather than the full expert set.
Specifically, for each token, KL-Top is applied to the union of the top-$k$ experts and an additional $\alpha (n-k)$ experts ranked immediately after the top-$k$, where $n$ is the total number of experts and $\alpha \in [0,1]$ is a hyperparameter controlling the extent of relaxation.
The probability distributions are computed as:
\begin{align}
p_{\text{fp}} &= \text{softmax}(\mathbf{\tilde x}\mathbf{\tilde W}^{\text{gate}}), \\
p_{\text{quant}} &= \text{softmax}(\mathbf{\tilde x}\mathcal{Q}(\mathbf{\tilde W}^{\text{gate}})).
\end{align}
The combined objective enforces both numerical fidelity and distributional alignment.
The MSE term stabilizes logit magnitudes, while the KL divergence explicitly constrains the routing probability simplex, ensuring that the quantized router preserves expert selection boundaries.
As a result, EAQuant effectively mitigates quantization-induced routing errors and maintains consistent expert activation patterns even under aggressive low-bit quantization.

\subsection{Expert-Aware Calibration Data Balance}
\label{subsec:data_balance}
To address the inherent expert activation imbalance in MoE models during post-training quantization, we propose an expert-aware calibration data balance strategy that explicitly leverages routing statistics to adaptively rebalance calibration samples across experts.
This strategy uses routing statistics to identify under-activated experts and applies threshold-driven oversampling until their activation counts reach the expected level, thereby reducing calibration bias in quantization parameter estimation.

Following standard PTQ practice, we first sample 128 sequences of length 4096 from WikiText2 to construct the base dataset $\mathcal{D}_{\text{base}}$. 
This data calibrates non-expert components such as the QKV layer and the gating network.
For MoE modules, we first forward $N=128 \times 4096$ tokens from $\mathcal{D}_{\text{base}}$ through the top-k router to obtain the profiling for token-expert assignment. For experts whose routed token counts fall below the expected activation level of ${r}\frac{kN}{n}$(e.g., the magnification ratio ${r}=2.0$) tokens, we iteratively sample new batches from the training dataset to construct $\mathcal{D}_{\text{expert}}$, until the routed tokens for these experts all surpass the average level of $r\frac{kN}{n}$ tokens.
Finally we use tokens from $\mathcal{D}_{\text{base}} \cup \mathcal{D}_{\text{expert}}$ to calibrate the quantization parameters for weights of local experts.

\section{Experiment}

\textbf{Models and Evaluations.} 
We perform comprehensive experiments across three SOTA MoE language models: DeepSeek-MoE-16B\footnote{https://huggingface.co/deepseek-ai/deepseek-moe-16b-base}~\cite{dai2024deepseekmoe}, OLMoE-7B\footnote{https://huggingface.co/allenai/OLMoE-1B-7B-0924}~\cite{muennighoff2025olmoe} and Mixtral-8x7B\footnote{https://huggingface.co/mistralai/Mixtral-8x7B-v0.1}~\cite{jiang2024mixtralexperts}. 
Beyond conventional perplexity evaluation on the Wikitext2~\cite{merity2016Wikitext2} and C4~\cite{raffel2023c4} benchmarks, 
We evaluate the proposed EAQuant on commonsense QA tasks via zero-shot accuracy across four challenging datasets: PIQA~\cite{bisk2019piqa}, ARC~\cite{clark2018arc}, BoolQ~\cite{clark2019boolq} and WinoGrande~(WG)~\cite{sakaguchi2019winogrande}.

\textbf{Our Baseline Method.} 
We adopt the SOTA post-training quantization (PTQ) method DuQuant~\cite{lin2024duquant} as our baseline.
For calibration, we follow the standard evaluation protocol and use 128 sequential text segments sampled from WikiText2.
The corresponding half-precision (FP16) model results are retained as reference baselines for fair and consistent performance comparison.

\textbf{Implementation Details.}
We set sequence length to 2048 for all evaluation tasks.
We apply asymmetric quantization with per-channel weight and per-token activation, without using group-wise quantization.
As an effective post-training quantization (PTQ) approach, our proposed EAQuant bypasses the need for parameter-sensitive fine-tuning.
We adapt the official repository of DuQuant to support the three MoE models.

\begin{table*}[t]
\centering
\small
\setlength{\tabcolsep}{3pt}
\renewcommand\arraystretch{1.1}
\caption{Results of DuQuant and EAQuant under various low-bit quantization settings on OLMoE-7B, DeepSeek-MoE-16B, and Mixtral-8x7B model. The router layer is quantized with W8A8.}
\vspace{-0.5em}
\label{tab:table1}
\begin{tabular}{l c c cc cccccc}
\toprule
\multirow{2}{*}{\textbf{Model}} &
\multirow{2}{*}{\textbf{Method}} &
\multirow{2}{*}{\makecell{\textbf{\#Bits}\\\textbf{(W-A)}}} &
\multicolumn{2}{c}{\textbf{PPL}$\downarrow$} &
\multicolumn{6}{c}{\textbf{Accuracy}$\uparrow$} \\  \cmidrule(r){4-5} \cmidrule(l){6-11}
& & & \textbf{WikiText2} & \textbf{C4} &
\textbf{ARC-C} & \textbf{ARC-E} & \textbf{BoolQ} & \textbf{PIQA} & \textbf{WG} & \textbf{Avg.} \\
\midrule

\multirow{9}{*}{\texttt{OLMoE-7B}}
& FP & 16-16 & 7.49 & 10.52 & 48.46 & 78.45 & 74.56 & 80.69 & 69.22 & 70.28 \\
\cline{2-11}

& DuQuant & 4-4 & 8.64 & 11.51 & 46.25 & 74.83 & 71.28 & 77.53 & 63.54 & 66.69 \\
& EAQuant & 4-4 & 8.52 & 11.41 & 47.35 & 76.14 & 72.29 & 78.84 & 65.67 &
\textbf{68.06}~(\textcolor{red}{\textbf{+1.37}}) \\
\cdashline{2-11}

& DuQuant & 3-4 & 10.77 & 13.59 & 41.64 & 70.71 & 65.75 & 75.35 & 63.06 & 63.30 \\
& EAQuant & 3-4 & 10.41 & 13.20 & 44.80 & 73.53 & 69.27 & 76.61 & 63.69 &
\textbf{65.58}~(\textcolor{red}{\textbf{+2.28}}) \\
\cdashline{2-11}

& DuQuant & 3-3 & 16.88 & 19.23 & 34.64 & 61.20 & 57.22 & 67.79 & 56.12 & 55.39 \\
& EAQuant & 3-3 & 14.14 & 16.68 & 37.71 & 67.13 & 58.90 & 69.64 & 58.41 &
\textbf{58.36}~(\textcolor{red}{\textbf{+2.97}}) \\
\cdashline{2-11}

& DuQuant & 2-4 & 8279.52 & 5508.23 & 24.15 & 34.01 & 41.19 & 55.28 & 51.46 & 41.22 \\
& EAQuant & 2-4 & 78.57 & 62.56 & 29.86 & 53.66 & 51.68 & 65.51 & 53.67 &
\textbf{50.88}~(\textcolor{red}{\textbf{+9.66}}) \\

\hline

\multirow{9}{*}{\texttt{\makecell{DeepSeek-MoE-16B}}}
& FP & 16-16 & 6.51 & 9.10 & 47.70 & 75.88 & 72.81 & 80.41 & 70.40 & 69.44 \\
\cline{2-11}

& DuQuant & 4-4 & 7.10 & 9.85 & 43.52 & 72.43 & 70.80 & 76.93 & 65.67 & 65.87 \\
& EAQuant & 4-4 & 7.06 & 9.78 & 44.03 & 72.90 & 71.35 & 78.89 & 67.96 &
\textbf{67.02}~(\textcolor{red}{\textbf{+1.15}}) \\
\cdashline{2-11}

& DuQuant & 3-4 & 8.28 & 11.41 & 40.44 & 70.20 & 65.23 & 76.22 & 61.64 & 62.75 \\
& EAQuant & 3-4 & 8.19 & 11.14 & 42.49 & 71.17 & 65.17 & 75.35 & 66.22 &
\textbf{64.08}~(\textcolor{red}{\textbf{+1.33}}) \\
\cdashline{2-11}

& DuQuant & 3-3 & 14.02 & 18.63 & 31.66 & 57.95 & 62.54 & 66.81 & 57.06 & 55.20 \\
& EAQuant & 3-3 & 11.62 & 15.31 & 35.15 & 60.82 & 61.68 & 71.44 & 57.62 &
\textbf{57.34}~(\textcolor{red}{\textbf{+2.14}}) \\
\cdashline{2-11}

& DuQuant & 2-4 & 10006.75 & 4114.85 & 23.98 & 29.38 & 39.79 & 53.43 & 50.43 & 39.40 \\
& EAQuant & 2-4 & 138.27 & 82.51 & 25.94 & 45.12 & 55.99 & 64.58 & 54.62 &
\textbf{49.25}~(\textcolor{red}{\textbf{+9.85}}) \\

\hline

\multirow{9}{*}{\texttt{\makecell{Mixtral-8x7B}}}
& FP & 16-16 & 3.84 & 6.98 & 55.80 & 83.29 & 84.56 & 83.41 & 75.85 & 76.58 \\
\cline{2-11}

& DuQuant & 4-4 & 4.47 & 7.47 & 52.22 & 79.21 & 81.25 & 80.74 & 71.90 & 73.06 \\
& EAQuant & 4-4 & 4.44 & 7.42 & 52.13 & 81.73 & 82.23 & 81.07 & 73.88 &
\textbf{74.21}~(\textcolor{red}{\textbf{+1.15}}) \\
\cdashline{2-11}

& DuQuant & 3-4 & 5.43 & 8.51 & 47.35 & 75.38 & 77.55 & 78.62 & 66.77 & 69.14 \\
& EAQuant & 3-4 & 5.27 & 8.23 & 50.68 & 78.45 & 78.69 & 79.05 & 69.30 &
\textbf{71.23}~(\textcolor{red}{\textbf{+2.09}}) \\
\cdashline{2-11}

& DuQuant & 3-3 & 11.18 & 16.47 & 32.94 & 59.47 & 53.06 & 66.10 & 53.67 & 53.05 \\
& EAQuant & 3-3 & 7.46 & 10.98 & 40.87 & 67.21 & 67.58 & 74.76 & 61.88 &
\textbf{62.46}~(\textcolor{red}{\textbf{+9.41}}) \\
\cdashline{2-11}

& DuQuant & 2-4 & 7626.38 & 6822.75 & 26.71 & 25.34 & 44.37 & 52.29 & 49.25 & 39.59 \\
& EAQuant & 2-4 & 21.94 & 25.24 & 32.00 & 57.15 & 55.90 & 67.46 & 54.46 &
\textbf{53.40}~(\textcolor{red}{\textbf{+13.81}}) \\

\bottomrule
\end{tabular}
\vspace{-0.5em}
\end{table*}

\begin{table*}[!tbp]
\centering
\small
\setlength\tabcolsep{1.5pt}
\renewcommand\arraystretch{1.1}
\caption{Ablation results of EAQuant components under the \textbf{W3A4} weight–activation quantization setting. The router layer is quantized with W8A8. EA-SA, EA-RCA, and EA-CDB represent Expert-Aware Smoothing Aggregation, Expert-Aware Routing Consistency Alignment, and Expert-Aware Calibration Data Balance, respectively.}
\vspace{-0.5em}
\label{tab:table3_ablation_module}
\begin{tabular}{ccc|cccccc|cccccc}
\hline
\multicolumn{3}{c|}{\textbf{Modules}} &
\multicolumn{6}{c|}{\textbf{OLMoE-7B}} &
\multicolumn{6}{c}{\textbf{DeepSeek-MoE-16B}} \\
\hline
\textbf{EA-SA} & \textbf{EA-RCA} & \textbf{EA-CDB} &
\textbf{ARC-C} & \textbf{ARC-E} & \textbf{BoolQ} & \textbf{PIQA} & \textbf{WG} & \textbf{Avg.} $\uparrow$&
\textbf{ARC-C} & \textbf{ARC-E} & \textbf{BoolQ} & \textbf{PIQA} & \textbf{WG} & \textbf{Avg.} $\uparrow$\\
\hline
 &  &  &
41.64 & 70.71 & 65.75 & 75.35 & 63.06 & 63.30 &
40.44 & 70.20 & 65.23 & 76.22 & 61.64 & 62.75 \\

\checkmark &  &  &
43.43 & \textbf{74.20} & 68.62 & \textbf{76.93} & 63.93 & 65.42~(\textcolor{red}{{+2.12}}) &
40.70 & 69.70 & \textbf{68.38} & \textbf{76.82} & 62.75 & 63.67~(\textcolor{red}{{+0.92}}) \\

 & \checkmark &  &
41.64 & 71.76 & 69.08 & 75.52 & 62.98 & 64.20~(\textcolor{red}{{+1.90}}) &
41.89 & 69.82 & 63.55 & 76.01 & 64.48 & 63.15~(\textcolor{red}{{+0.40}}) \\

 &  & \checkmark &
40.27 & 70.16 & 67.71 & 75.84 & \textbf{64.25} & 63.65~(\textcolor{red}{{+0.35}}) &
40.10 & 69.53 & 64.59 & 76.06 & 64.01 & 62.86~(\textcolor{red}{{+0.11}}) \\

\checkmark & \checkmark &  &
44.69 & 73.57 & 68.93 & 76.71 & 63.54 & 65.49~(\textcolor{red}{{+2.19}}) &
41.61 & 70.81 & 64.93 & 76.28 & 65.43 & 63.81~(\textcolor{red}{{+1.06}}) \\

\checkmark & \checkmark & \checkmark &
\textbf{44.80} & {73.53} & \textbf{69.27} & {76.61} & {63.69} & \textbf{65.58}~(\textcolor{red}{\textbf{+2.28}}) &
\textbf{42.49} & \textbf{71.17} & {65.17} & {75.35} & \textbf{66.22} & \textbf{64.08}~(\textcolor{red}{\textbf{+1.33}}) \\
\hline
\end{tabular}
\vspace{-1em} \end{table*}

\begin{table*}[t]
    \centering
    \small
    \setlength\tabcolsep{2.0pt}
    \renewcommand\arraystretch{1.1}
    \caption{Ablation of the \textbf{EA-SA} smooth aggregation strategy $\mathbb{A}$ under the \textbf{W4A4} weight-activation quantization setting. We compare different expert-aware aggregation designs, including \textbf{maximum}, \textbf{expert\_frequency}, and \textbf{router\_logits}, with DuQuant as the baseline.}

    \vspace{-0.5em}
    \label{tab:table7_ablation_smooth_strategy}
    
    \begin{tabular}{llcccccccc}
    \toprule
    \multirow{2}*{\textbf{Model}} & \multirow{2}*{\textbf{Method}} 
    & \multicolumn{2}{c}{\textbf{PPL}$\downarrow$} 
    & \multicolumn{6}{c}{\textbf{ACCURACY}$\uparrow$}  \\
    \cmidrule(r){3-4} \cmidrule(l){5-10}
    & & \textbf{WikiText2} & \textbf{C4} 
    & \textbf{ARC-C} & \textbf{ARC-E} & \textbf{BoolQ} 
    & \textbf{PIQA} & \textbf{WinoGrande} & \textbf{Avg.}  \\
     \midrule

    \multirow{5}*{\texttt{OLMoE-7B}} 
    & FP            & 7.49  & 10.52 & 48.46 & 78.45 & 74.56 & 80.69 & 69.22 & 70.28 \\ 
    \cline{2-10}
    & Baseline   & 8.64  & 11.51 & 46.25 & 74.83 & 71.28 & 77.53 & 63.54 & 66.69 \\ 
    & + maximum        & \textbf{8.54}  & 11.43 & \textbf{47.44} & 75.04 & \textbf{72.45} & \textbf{79.27} & 65.27 & \textbf{67.89}~(\textcolor{red}{\textbf{+1.20}}) \\
    & + expert\_frequency & 8.56  & 11.46 & 45.14 & \textbf{76.09} & 71.38 & 78.94 & \textbf{66.06} & 67.52~(\textcolor{red}{{+0.83}}) \\
    & + router\_logits  & 8.57  & 11.47 & 45.99 & 75.63 & 71.44 & 78.45 & 65.51 & 67.40~(\textcolor{red}{{+0.71}}) \\
     \hline

    \multirow{5}*{\texttt{DeepSeek-MoE-16B}} 
    & FP            & 6.51 & 9.10 & 47.70 & 75.88 & 72.81 & 80.41 & 70.40 & 69.44 \\ 
    \cline{2-10}
    & Baseline   & 7.10 & 9.85 & 43.52 & 72.43 & 70.80 & 76.93 & 65.67 & 65.87 \\ 
    & + maximum        & 7.09 & 9.83 & 44.03 & 73.44 & \textbf{69.57} & \textbf{78.51} & 65.90 & 66.29~(\textcolor{red}{{+0.42}}) \\
    & + expert\_frequency & 7.08 & 9.81 & 43.17 & 73.19 & 69.39 & 78.07 & 68.51 & 66.47~(\textcolor{red}{{+0.60}}) \\
    & + router\_logits  & \textbf{7.07} & \textbf{9.80} & \textbf{44.58} & \textbf{73.52} & 69.23 & 78.24 & \textbf{69.02} & \textbf{66.92}~(\textcolor{red}{\textbf{+1.05}}) \\
    \hline

    \multirow{5}*{\texttt{Mixtral-8x7B}} 
    & FP            & 3.84 & 6.98 & 55.80 & 83.29 & 84.56 & 83.41 & 75.85 & 76.58 \\ 
    \cline{2-10}
    & Baseline  & 4.47 & 7.47 & 52.22 & 79.21 & 81.25 & 80.74 & 71.90 & 73.06 \\ 
    & + maximum        & 4.47 & 7.46 & 53.06 & 80.02 & 82.21 & 81.51 & 70.81 & 73.52~(\textcolor{red}{{+0.46}}) \\
    & + expert\_frequency & \textbf{4.44} & \textbf{7.43} & 53.29 & \textbf{80.43} & \textbf{82.63} & 81.07 & \textbf{72.49} & \textbf{73.98}~(\textcolor{red}{\textbf{+0.92}}) \\
    & + router\_logits  & 4.45 & 7.45 & \textbf{53.58} & 79.67 & 82.42 & \textbf{81.18} & 71.58 & 73.69~(\textcolor{red}{{+0.63}}) \\
    \bottomrule
    \end{tabular}
\vspace{-0.5em}
\end{table*}

\begin{table*}[!t]
    \centering
    \small
    \renewcommand\arraystretch{1.1}
    \caption{Ablation of \textbf{EA-RCA} with different weight-activation quantization configuration among 7 tasks on OLMoE-7B. Notably, Rw*a* represents the weight-activation quantization configuration of router layer.  DuQuant serves as the baseline method.}
    \vspace{-0.5em}
    \label{tab:table4_ablation_router}
    \begin{tabular}{llcccccccc}
    \toprule
    \multirow{2}*{\textbf{Model}} & \multirow{2}*{\textbf{Method}} & \multicolumn{2}{c}{\textbf{PPL}$\downarrow$} & \multicolumn{6}{c}{\textbf{ACCURACY}$\uparrow$}  \\
    \cmidrule(r){3-4} \cmidrule(l){5-10}
    & & \textbf{WikiText2} & \textbf{C4} & \textbf{ARC-C} & \textbf{ARC-E} & \textbf{BoolQ} & \textbf{PIQA} & \textbf{WG} & \textbf{Avg.} $\uparrow$ \\
    \midrule
    \multirow{9}*{\texttt{OLMoE-7B}} 
    & FP & 7.49 & 10.52 & 48.46 & 78.45 & 74.56 & 80.69 & 69.22 & 70.28 \\
    \cline{2-10}
    & w3a4\_Rw3a4\_Baseline & 11.09 & 13.95 & 40.70 & 71.84 & 62.69 & 75.24 & 61.17 & 62.33 \\
    & {+ EA-RCA} & 11.07 & 13.94 & 40.87 & 71.51 & 67.55 & 75.73 & 62.51 & \textbf{63.63}~(\textcolor{red}{\textbf{+1.30}}) \\
    \cline{2-10}
    & w3a4\_Rw4a4\_Baseline & 10.94 & 13.68 & 40.70 & 71.25 & 64.98 & 75.63 & 62.43 & 63.00 \\
    & { + EA-RCA} & 10.85 & 13.67 & 40.96 & 71.55 & 68.04 & 75.30 & 64.09 & \textbf{63.99}~(\textcolor{red}{\textbf{+0.99}}) \\
    \cline{2-10}
    & w3a4\_Rw8a8\_Baseline & 10.77 & 13.59 & 41.64 & 70.71 & 65.75 & 75.35 & 63.06 & 63.30 \\
    & { + EA-RCA} & 10.71 & 13.56 & 41.64 & 71.76 & 69.08 & 75.52 & 62.98 & \textbf{64.20}~(\textcolor{red}{\textbf{+0.90}}) \\
    \cline{2-10}
    & w4a4\_Rw8a8\_Baseline & 8.64 & 11.51 & 46.25 & 74.83 & 71.28 & 77.53 & 63.54 & 66.69 \\
    & { + EA-RCA} & 8.60 & 11.48 & 46.76 & 75.97 & 72.14 & 77.80 & 67.17 & \textbf{67.97}~(\textcolor{red}{\textbf{+1.28}}) \\
    \bottomrule
    \end{tabular}
    \vspace{-1.5em}
 \end{table*}

\begin{table}[t]
    \centering
    \small
    \setlength\tabcolsep{6.0pt}
    \renewcommand\arraystretch{1.1}
    \caption{Ablation of the magnification ratio $r$ in \textbf{EA-CDB} with \textbf{W4A4} weight-activation quantization configuration on OLMoE model. Notably, the router layer is quantized with W8A8.}
    \vspace{-0.5em}
    \label{tab:table6_ablation_calib_r}
    \begin{tabular}{c*{5}{c}}
        \toprule
        $r$ & 0.0 & 1.0 & 2.0 & 4.0 & 8.0 \\
        \midrule
        Avg.$\uparrow$         & 66.69 & 67.10 & \textbf{67.78} & 67.47 & 67.32 \\
        \bottomrule
    \end{tabular}
    \vspace{-1.5em} 
\end{table}

\subsection{Main Results}
\label{subsec:Main_Results}
\textbf{Comparison Results.}  
Table~\ref{tab:table1} summarizes quantization results on OLMoE-7B, DeepSeek-MoE-16B, and Mixtral-8x7B. EAQuant consistently outperforms DuQuant across all evaluated settings, demonstrating strong robustness under low-bit quantization.
Under the W4A4 configuration, EAQuant achieves average accuracy improvements of 1.37\%, 1.15\%, and 1.15\% across the three models, with notable gains on reasoning benchmarks (e.g., +2.52\% on ARC-E for Mixtral-8x7B) and better perplexity alignment with full-precision models. As quantization becomes more aggressive (e.g., W3A4), EAQuant further enlarges the performance gap.
More importantly, under extreme low-bit settings such as \textbf{W3A3} and \textbf{W2A4}, existing PTQ methods suffer from severe performance degradation or numerical instability, especially on large-scale MoE models. In contrast, EAQuant maintains stable perplexity and preserves strong reasoning accuracy, achieving substantial gains over DuQuant, with improvements reaching up to \textbf{+13.81\%} average accuracy on Mixtral-8x7B. These results indicate that EAQuant is particularly effective in ultra-low-bit, in which preserving expert routing behavior and balanced expert calibration becomes critical for reliable MoE quantization.

\subsection{Ablation Study}
\label{subsec:Ablation_Study}

\textbf{Module-wise Impact.}
We perform a module-wise ablation study under the W3A4 quantization setting to analyze the contribution of each component in EAQuant. As shown in Table~\ref{tab:table3_ablation_module}, all modules positively contribute to quantized MoE performance, with Expert-Aware Smoothing Aggregation (EA-SA) providing the largest individual gains by effectively mitigating activation outliers.
Expert-Aware Routing Consistency Alignment (EA-RCA) yields additional improvements by stabilizing expert routing, while Expert-Aware Calibration Data Balance (EA-CDB) offers complementary benefits by alleviating calibration bias. Combining EA-SA and EA-RCA already outperforms any single module, and the full EAQuant framework achieves the best results, with average accuracy improvements of +2.28 on OLMoE-7B and +1.33 on DeepSeek-MoE-16B. These results demonstrate that the three components are complementary and jointly enable robust low-bit MoE quantization.

\textbf{Ablation Analysis of EA-SA.}
Table~\ref{tab:table7_ablation_smooth_strategy} presents an ablation study of the smooth aggregation strategy in EA-SA, highlighting its importance in mitigating post-training quantization (PTQ) degradation for MoE models. Compared to the DuQuant baseline, which exhibits notable drops in both perplexity and accuracy, all three expert-aware aggregation strategy, \textbf{maximum}, \textbf{expert\_frequency}, and \textbf{router\_logits}, consistently recover quantized performance.
Specifically, \textbf{maximum} aggregates experts via max-scaling across weights, \textbf{expert\_frequency} applies activation-frequency-based weighting, and \textbf{router\_logits} leverages routing probabilities. Among them, \textbf{maximum} achieves the strongest overall improvement (+1.20 average accuracy), indicating superior preservation of dominant expert signals. In contrast, \textbf{expert\_frequency} and \textbf{router\_logits} exhibit complementary gains on tasks such as ARC-E and WinoGrande, underscoring the value of incorporating expert utilization and routing dynamics.
Overall, these results confirm that task-aware, expert-aware aggregation is crucial for addressing quantization-induced activation irregularities while preserving MoE performance across benchmarks.

\textbf{Effectiveness of EA-RCA.}
We evaluate the impact of Expert-Aware Routing Consistency Alignment (EA-RCA) on OLMoE-7B under various weight-activation quantization settings. As shown in Table~\ref{tab:table4_ablation_router}, removing EA-RCA consistently degrades performance across most tasks, with more severe drops under low-bit quantization. Under W3A4, disabling EA-RCA reduces average accuracy by up to 1.30 points and harms routing-sensitive tasks like BoolQ and WinoGrande, reflecting increased routing instability. This trend persists across different router quantization configurations (Rw3a4, Rw4a4, Rw8a8), where EA-RCA improves average accuracy by 0.9–1.3 points. These results confirm EA-RCA’s effectiveness in stabilizing routing distributions and mitigating quantization-induced inconsistencies, particularly under aggressive quantization. Further ablations on KL-Top are provided in the Appendix.

\textbf{Effectiveness of EA-CDB.}
We investigate the effect of the magnification ratio $r$ in the calibration data balance (EA-CDB) module on OLMoE-7B. The ratio $r$ controls the minimum token threshold for expert calibration, encouraging underutilized experts to receive sufficient calibration data and alleviating activation imbalance during quantization.
As shown in Table~\ref{tab:table6_ablation_calib_r}, setting $r=2.0$ yields the best average accuracy across tasks. In contrast, disabling calibration balance ($r=0.0$) results in the lowest performance, highlighting the importance of balanced expert-aware calibration for mitigating quantization-induced errors.
Table~\ref{tab:layer15_quant_error} shows expert-level quantization errors for down-projection weights in layer 15. EA-CDB reduces the average error by 19.4\% and lowers it for 87.5\% of experts, while suppressing extreme outliers by reducing maximum and minimum errors. This ensures robustness improvements across all experts, not just a subset.

\begin{table}[!t]
\centering
\caption{Quantization error statistics of OLMoE Layer~15 experts.
\# RE denotes the number of experts whose quantization error is reduced when EA-CDB is applied. The complete per-expert quantization error results are provided in the Appendix.}
\vspace{-0.5em}
    \renewcommand\arraystretch{1.1}
\label{tab:layer15_quant_error}
\begin{tabular}{lccc}
\toprule
\textbf{Metric} & \textbf{w/o EA-CDB} & \textbf{w/ EA-CDB} & \textbf{Reduction} \\
\midrule
Mean     & 0.0144 & 0.0116 & $\downarrow$ 19.4\% \\
Median   & 0.0116 & 0.0098 & $\downarrow$ 15.5\% \\
Max      & 0.0381 & 0.0361 & $\downarrow$ 5.2\%  \\
Min      & 0.0069 & 0.0032 & $\downarrow$ 53.6\% \\
Std      & 0.0070 & 0.0064 & $\downarrow$ 8.6\%  \\
\midrule
\# RE & -- & 56 / 64 & \textbf{87.5}\% \\
\bottomrule
\end{tabular}
    \vspace{-1.5em}
\end{table}

\subsection{Comparison with MoE Quantization Methods}
Table~\ref{tab:moe_quant_compare} compares EAQuant with representative MoE quantization methods on OLMoE-7B, DeepSeek-MoE-16B, and Mixtral-8x7B under different weight-activation precision settings. Across both moderate (4-16) and aggressive (3-16) configurations, EAQuant consistently achieves the lowest perplexity on WikiText2 and C4, demonstrating strong robustness under low-bit quantization.
Notably, several existing methods suffer severe performance degradation or  numerical instability at 3-bit weight precision; for example, MoEQuant-AWQ diverges on DeepSeek-MoE-16B under the 3-16 setting, whereas EAQuant remains stable and delivers substantially lower perplexity. On Mixtral-8x7B, EAQuant further shows good scalability across precision regimes, outperforming competing methods

\begin{table}[!t]
\centering
\small
\setlength{\tabcolsep}{1.4pt}
\renewcommand{\arraystretch}{1.1}
\caption{Comparison with existing MoE quantization methods.}
\vspace{-0.5em}
\label{tab:moe_quant_compare}
    \renewcommand\arraystretch{1.1}
\begin{tabular}{cc c cc}
\toprule
\textbf{Model} & \textbf{Method} & \textbf{W-A} &
\textbf{WikiText2}$\downarrow$ & \textbf{C4}$\downarrow$ \\
\midrule
\multirow{4}{*}{\texttt{\makecell{OLMoE-7B}}}
& GPTQ&3-16&11.65&18.86\\
& MoEPTQ&3.26-16&15.44&26.04\\
& MoQa&2.95-16&10.59&18.42\\
& \textbf{EAQuant (Ours)}&3-16&\textbf{9.60}	&\textbf{12.38}
\\ \hline

\multirow{9}{*}{\texttt{\makecell{DeepSeek-\\MoE-16B}}}
& MoEQuant-AWQ & 4-16 & 6.94 & 9.32 \\
& MoEQuant-GPTQ & 4-16 & 6.78 & 9.22 \\
& \textbf{EAQuant (Ours)} & 4-16 &
\textbf{6.70} & \textbf{9.18} \\
\cdashline{2-5}

& GPTQ & 3-16 & 10.47 & 15.19 \\
& MoEPTQ & 3.31-16 & 8.49 & 15.61 \\
& MoQa & 2.98-16 & 7.94 & 13.49 \\
& MoEQuant-AWQ & 3-16 & 5100 & 4924 \\
& MoEQuant-GPTQ & 3-16 & 7.55 & 10.88 \\
& \textbf{EAQuant (Ours)} & 3-16 &
\textbf{7.27} & \textbf{10.04} \\
\hline
\multirow{10}{*}{\texttt{\makecell{Mixtral-\\8x7B}}}
& MoEQuant-AWQ & 4-16 & 5.15 & 7.84 \\
& MoEQuant-GPTQ & 4-16 & 4.12 & 7.34 \\
& \textbf{EAQuant (Ours)} & 4-16 &
\textbf{4.03} & \textbf{7.10} \\
\cdashline{2-5}

& MoEQuant-AWQ & 3-16 & 8.77 & 11.44 \\
& MoEQuant-GPTQ & 3-16 & 4.90 & 8.24 \\
& \textbf{EAQuant (Ours)} & 3-16 &
\textbf{4.49} & \textbf{7.47} \\
\cdashline{2-5}

& MCMoE & 2.2-16 & 5.72 & -- \\
& MCMoE & 2.05-16 & 5.91 & -- \\
& \textbf{EAQuant (Ours)} & 2-16 &
\textbf{5.65} & 8.56 \\
\bottomrule
\end{tabular}
    \vspace{-2em} 
\end{table}

\section{Conclusion}

This work presents EAQuant, an expert-aware post-training quantization framework for Mixture-of-Experts models, which addresses activation outliers, routing instability, and expert sparsity via expert-aware smoothing aggregation, routing consistency alignment, and calibration data balancing. 
Extensive experiments on OLMoE-7B, DeepSeek-MoE-16B, and Mixtral-8x7B show that EAQuant consistently outperforms existing PTQ methods, achieving 1.15-1.37\% average accuracy improvements under W4A4 and robust 1.33-2.28\% gains under W3A4. Under more aggressive settings such as W3A3 and W2A4, EAQuant substantially mitigates performance collapse and outperforms prior MoE quantization methods with up to 13.81\% gains. These results demonstrate that preserving expert-aware activation scaling, routing behavior, and proper calibration under sparse expert activation is critical for reliable low-bit quantization of MoE models.

\section*{Limitation and Future Work}
While EAQuant improves PTQ for MoE models under extreme quantization (e.g., W4A4/W3A4/W3A3/W2A4), its performance degrades in ultra-low-bit regimes (e.g., W1A4) due to the granularity-sparsity trade-off, requiring QAT for accuracy recovery. However, QAT’s high computational cost and training data dependency hinder deployment in resource-constrained settings. Future work may explore hybrid PTQ-QAT approaches or expert-specific calibration to balance efficiency and performance.

\section*{Impact Statement}
This work proposes EAQuant, an expert-aware optimization framework that addresses key challenges in post-training quantization for Mixture-of-Experts (MoE) models. By reducing accuracy loss during quantization, EAQuant enables efficient deployment of MoE models in resource-constrained settings, lowering computational and energy costs for applications like NLP and edge computing.

\bibliography{example_paper}
\bibliographystyle{icml2026}

\appendix
\clearpage
\onecolumn

\section*{Appendix Overview}
\begin{itemize}
    \item Section~\ref{sec:appx_related_work}: Related work.
    \item Section~\ref{sec:appx_experimental}: More experimental results.
    \item Section~\ref{sec:appx_vis}: More visualization examples.
\end{itemize}

\section{Related Work}
\label{sec:appx_related_work}

\subsection{Mixture-of-Experts Large Language Models}
The Mixture-of-Experts (MoE) paradigm, originally proposed by~\cite{jacobs1991adaptive} and~\cite{jordan1994hierarchical}, has emerged as a pivotal architecture for scaling Large Language Models (LLMs). In contemporary MoE LLMs, each layer comprises multiple expert networks alongside a gating network, typically implemented as a linear layer with softmax activation, that dynamically routes inputs to selected experts and aggregates their outputs. 
Various architectural innovations have been proposed to optimize this routing mechanism and expert configuration. Switch Transformer ~\cite{fedus2022switch} popularized the top-1 gating strategy, demonstrating competitive performance at scale, while Mixtral-8x7B~\cite{jiang2024mixtralexperts} achieved an effective balance between performance and computational efficiency by activating merely two experts per layer. More recently, DeepSeekMoE~\cite{dai2024deepseekmoe} introduced fine-grained expert segmentation through the subdivision of FFN hidden dimensions, accompanied by shared experts that remain constantly activated to capture common knowledge and reduce parameter redundancy,
an approach further refined in subsequent iterations DeepSeek-V2~\cite{liu2024deepseek} and DeepSeek-V3~\cite{liu2024deepseekv3}.
Concurrent architectural explorations include OLMoE~\cite{muennighoff2025olmoe}, which advances open-weight MoE models with sophisticated training strategies, as well as Qwen-MoE~\cite{qwen15moe}, which aggressively replaces traditional FFN layers entirely with MoE layers to enhance pretraining performance through progressive adaptation from dense model checkpoints.

\subsection{Post-Training Quantization for LLMs}
Modern large language models predominantly rely on the Transformer architecture, a design that demands considerable memory resources during both training and inference. To mitigate this limitation, Post-training quantization (PTQ) has emerged as a mainstream compression paradigm, enabling substantial reductions in memory footprint while preserving the model's predictive performance. 
Among prominent PTQ approaches, GPTQ~\cite{frantar2023gptq} employs Hessian-based error compensation to achieve high compression rates, 
while AWQ~\cite{lin2023awq} considers activation distributions to improve weight quantization. 
Beyond weight-only quantization, addressing activation outliers has proven crucial for effective quantization. 
SmoothQuant~\cite{xiao2023smoothquant} migrates quantization difficulty from activations to weights through channel-wise scaling, whereas Outlier Suppression+~\cite{wei2023osplus} introduces optimal shifting to suppress outliers without fine-tuning. 
Recent advances further push the boundaries of low-bit quantization, including DuQuant~\cite{lin2024duquant} with Hadamard-based distribution-aware transformations, FlatQuant~\cite{sun2025flatquant} which flattens activation distributions through adaptive inter-channel transformations, and OstQuant~\cite{hu2025ostquant} that jointly optimizes weights and activations via outlier-aware associated subnets.

\subsection{Post-Training Quantization for MoEs}
Recent progress has been made in post-training quantization for Mixture-of-Experts (MoE).
MC-MoE~\cite{huang2025MCMOE} optimizes compression gains by exploiting structural redundancy in MoE models, while MoEQuant~\cite{hu2025moequant} introduces expert-balanced self-sampling to address inter- and intra-expert calibration imbalances. MoQE~\cite{kim2023moqe} explores the complementary relationship between low-bit quantization and model robustness, and QuantMoE-Bench~\cite{li2025quantmoebench} provides a systematic evaluation framework for quantization performance. MoQa~\cite{zheng2025moqa} further proposes a multi-stage data-model distribution-aware approach to align quantization strategies with expert-specific characteristics. 
However, these approaches primarily focus on weight or activation compression, leaving two critical challenges underexplored: activation distribution patterns in MoE systems and quantization sensitivity in routing layers. 
Notably, quantization-induced perturbations in routing logits cause structural instability and workload imbalance, while activation outliers in MoE systems exhibit channel-wise concentration and cross-expert correlations that remain unexplored in existing strategies.
This critical gap highlights the need for methods that directly target both activation distribution stability and routing layer robustness in low-precision MoE implementations.

\section{More Experimental Results}
\label{sec:appx_experimental}

This appendix provides additional experimental results and detailed analyses to further support the design choices of EAQuant. These results complement the main paper by offering deeper insights into the behavior of expert-aware routing consistency 
 alignment (EA-RCA) and expert-aware calibration data balance (EA-CDB) under low-bit post-training quantization.

\subsection{More Ablation Analysis of Expert-Aware Routing Consistency Alignment}

Table~\ref{tab:table5_ablation_router_KLtop} presents a detailed ablation study on the KL-Top loss used in the Expert-Aware Routing Consistency Alignment (EA-RCA) module under the W4A4 quantization setting, where the router layer is quantized with W8A8. This experiment investigates how restricting the KL divergence regularization to a subset of experts affects routing stability and downstream task performance.
Specifically, the KL-Top mechanism controls the number of experts involved in KL divergence computation by a ratio parameter $\alpha$, where $\alpha=0.00$ restricts the regularization to the top-$k$ experts selected by the router (with $k=8$ for OLMoE-7B), and $\alpha=1.00$ includes all experts. More generally, the number of experts $m$ involved in the KL term is given by
\begin{equation}
m = k + \lfloor \alpha \times (n - k) \rfloor,
\end{equation}
where $n$ denotes the total number of experts. As shown in Table~\ref{tab:table5_ablation_router_KLtop}, constraining the KL loss to only the top-$k$ experts ($\alpha=0.00$) yields the best overall performance, improving the average accuracy from 66.69 to 67.97 compared to the DuQuant baseline. This improvement is particularly pronounced on reasoning-oriented benchmarks such as ARC-E (+1.14) and WinoGrande (+3.63), while also slightly reducing perplexity on WikiText2 and C4. These results indicate that focusing KL regularization on high-confidence experts effectively preserves critical routing decision boundaries under quantization.

In contrast, gradually expanding the KL constraint to include lower-confidence experts (i.e., increasing $\alpha$) leads to diminished gains and, in some cases, performance degradation. This suggests that excessive regularization over weakly activated experts introduces noise into the routing distribution, which can interfere with the sparse expert selection mechanism. Overall, this ablation validates the necessity of selectively aligning routing distributions for high-impact experts rather than enforcing global distribution matching.

\begin{table*}[!h]
    \centering
    \small
    \renewcommand\arraystretch{1.1}
    \caption{Ablation of KL-Top loss in \textbf{EA-RCA} with \textbf{W4A4} weight-activation quantization configuration. Notably, the router layer is quantized with W8A8. DuQuant serves as the baseline method.}
    \label{tab:table5_ablation_router_KLtop}
    \begin{tabular}{llcccccccc}
    \toprule
    \multirow{2}*{\textbf{Model}} & \multirow{2}*{\textbf{Method}} & \multicolumn{2}{c}{\textbf{PPL}$\downarrow$} & \multicolumn{6}{c}{\textbf{ACCURACY}$\uparrow$}  \\
    \cmidrule(r){3-4} \cmidrule(l){5-10}
    & & \textbf{WikiText2} & \textbf{C4} & \textbf{ARC-C} & \textbf{ARC-E} & \textbf{BoolQ} & \textbf{PIQA} & \textbf{WG} & \textbf{Avg.} $\uparrow$ \\
    \midrule
    \multirow{7}*{\texttt{OLMoE-7B}} 
    & FP            & 7.49  & 10.52 & 48.46 & 78.45 & 74.56 & 80.69 & 69.22 & 70.28 \\ 
    \cline{2-10}
    & Baseline   & 8.64  & 11.51 & 46.25 & 74.83 & 71.28 & 77.53 & 63.54 & 66.69 \\ 
    & + KL-Top~($\alpha=0.00$)     & 8.60  & 11.48 & \textbf{46.76} & 75.97 & \textbf{72.14} & 77.80 & \textbf{67.17} & \textbf{67.97}~(\textcolor{red}{\textbf{+1.28}})  \\ 
    & + KL-Top~($\alpha=0.25$)    & 8.61  & 11.50 & 45.39 & 75.38 & 71.65 & \textbf{78.73} & 66.22 & 67.47~(\textcolor{red}{{+0.78}})  \\ 
    & + KL-Top~($\alpha=0.50$)    & 8.63  & 11.51 & 46.33 & 75.08 & 70.86 & 78.56 & 65.90 & 67.35~(\textcolor{red}{{+0.66}}) \\ 
    & + KL-Top~($\alpha=0.75$)    & 8.61  & 11.49 & 46.08 & \textbf{76.01} & 71.07 & 78.40 & 66.38 & 67.59~(\textcolor{red}{{+0.90}}) \\
    & + KL-Top~($\alpha=1.00$)   & 8.64  & 11.51 & 46.50 & 75.42 & 71.35 & 78.13 & 64.56 & 67.19~(\textcolor{red}{{+0.50}}) \\ 
    \bottomrule
    \end{tabular}
\vspace{-1em} \end{table*}

\subsection{Detailed Analysis of Expert-Aware Calibration Data Balance}

Table~\ref{tab:supp_ablation_calib_r_full} reports the full per-task performance breakdown for different magnification ratios $r$ used in the Expert-Aware Calibration Data Balance (EA-CDB) module under W4A4 quantization on OLMoE-7B. While the main paper reports only the averaged accuracy, this table provides finer-grained insights into how calibration balance affects individual tasks.
The magnification ratio $r$ controls the minimum number of routed tokens required for each expert during calibration. When $r=0$, no explicit balancing is applied, resulting in suboptimal performance due to severe expert-level data imbalance. Increasing $r$ improves performance across most tasks, with $r=2.0$ achieving the best average accuracy of 67.78. Notably, this setting consistently improves reasoning and commonsense tasks such as ARC-E, BoolQ, and PIQA, indicating more reliable quantization parameter estimation for underutilized experts.
Further increasing $r$ beyond 2.0 yields diminishing returns or slight performance degradation, suggesting that excessive oversampling may introduce distribution shift or redundant calibration data. These results empirically justify the default choice of $r=2.0$ used throughout our experiments.

\begin{table*}[!h]
    \centering
    \small
    \setlength{\tabcolsep}{4.5pt}
    \renewcommand{\arraystretch}{1.2}
    \caption{Detailed ablation results of the magnification ratio $r$ in \textbf{EA-CDB} under the \textbf{W4A4} weight-activation quantization setting on \texttt{OLMoE-7B}. The router layer is quantized with W8A8.}
    \label{tab:supp_ablation_calib_r_full}
    \begin{tabular}{c ccccc c}
        \toprule
        \textbf{$r$} 
        & \textbf{ARC-C} 
        & \textbf{ARC-E} 
        & \textbf{BoolQ} 
        & \textbf{PIQA} 
        & \textbf{WG} 
        & \textbf{Avg.} $\uparrow$ \\
        \midrule
        0.0 
        & 46.25 
        & 74.83 
        & 71.28 
        & 77.53 
        & 63.54 
        & 66.69 \\
        
        1.0 
        & 44.97 
        & 75.42 
        & 72.11 
        & 77.97 
        & 65.04 
        & 67.10 \\
        
        2.0 
        & \textbf{46.50} 
        & \textbf{76.01} 
        & \textbf{72.26} 
        & \textbf{78.24} 
        & \textbf{65.90} 
        & \textbf{67.78} \\
        
        4.0 
        & 45.95 
        & 75.97 
        & 71.97 
        & 78.18 
        & 65.29 
        & 67.47 \\
        
        8.0 
        & 46.08 
        & \textbf{76.01} 
        & 71.99 
        & 77.42 
        & 65.11 
        & 67.32 \\
        \bottomrule
    \end{tabular}
\vspace{-1em} \end{table*}

\subsection{Scaling to Larger MoE Models}

To evaluate the scalability of EAQuant, Table~\ref{tab:mixtral_8x22b_ppl} compares EAQuant with DuQuant on the large-scale Mixtral-8x22B\footnote{https://huggingface.co/mistralai/Mixtral-8x22B-v0.1} model. Despite the substantially increased number of parameters, EAQuant consistently achieves lower perplexity across all quantization settings. In particular, under aggressive low-bit configurations (e.g., W2A4), EAQuant dramatically mitigates catastrophic divergence observed in the baseline, reducing WikiText2 perplexity from 108.66 to 7.07. These results demonstrate that EAQuant scales robustly to large MoE architectures with huge parameters.

\begin{table}[!h]
\centering
\small
\setlength{\tabcolsep}{4pt}
\renewcommand{\arraystretch}{1.2}
\caption{Perplexity comparison of DuQuant and EAQuant on Mixtral-8x22B.}
\label{tab:mixtral_8x22b_ppl}
\begin{tabular}{l l c cc}
\toprule
\textbf{Model} & \textbf{Method} & \textbf{W-ACT} &
\textbf{WikiText2}$\downarrow$ & \textbf{C4}$\downarrow$ \\
\midrule

\multirow{7}{*}{\texttt{Mixtral-8x22B}}
& FP & 16-16 & 2.83 & 6.34 \\
\cline{2-5}

& DuQuant & 4-4 & 3.40 & 6.72 \\
& EAQuant & 4-4 &
\textbf{3.31}~(\textcolor{red}{\textbf{-0.09}}) &
\textbf{6.65}~(\textcolor{red}{\textbf{-0.07}}) \\
\cdashline{2-5}

& DuQuant & 3-3 & 6.20 & 9.95 \\
& EAQuant & 3-3 &
\textbf{5.39}~(\textcolor{red}{\textbf{-0.81}}) &
\textbf{8.70}~(\textcolor{red}{\textbf{-1.25}}) \\
\cdashline{2-5}

& DuQuant & 2-4 & 108.66 & 100.61 \\
& EAQuant & 2-4 &
\textbf{7.07}~(\textcolor{red}{\textbf{-101.59}}) &
\textbf{10.72}~(\textcolor{red}{\textbf{-89.89}}) \\

\bottomrule
\end{tabular}
    \vspace{-1em} \end{table}

\subsection{Impact of Calibration Dataset}

Table~\ref{tab:olmoe7b_calibration} studies the effect of using C4 instead of WikiText2 as the calibration dataset on OLMoE-7B. Across both W4A4 and W3A4 settings, EAQuant consistently outperforms DuQuant in terms of both perplexity and downstream accuracy. This indicates that EAQuant is not sensitive to the choice of calibration corpus and generalizes well across different data distributions.
\begin{table*}[!h]
\centering
\small
\setlength{\tabcolsep}{3pt}
\renewcommand{\arraystretch}{1.2}
\caption{Performance of DuQuant and EAQuant on OLMoE-7B with C4 as the calibration dataset.}
\label{tab:olmoe7b_calibration}
\begin{tabular}{l l c c cc cccccc}
\toprule
\multirow{2}*{\textbf{Model}} & \multirow{2}*{\textbf{Method}} & \multirow{2}*{\textbf{W-ACT}} & \multirow{2}*{\textbf{Calib.}} &
\multicolumn{2}{c}{\textbf{PPL}$\downarrow$} &
\multicolumn{6}{c}{\textbf{Accuracy}$\uparrow$} \\
\cmidrule(lr){5-6} \cmidrule(lr){7-12}
 & & & &
\textbf{WikiText2} & \textbf{C4} &
\textbf{ARC-C} & \textbf{ARC-E} & \textbf{BoolQ} &
\textbf{PIQA} & \textbf{WinoGrande} & \textbf{Avg.} \\
\midrule

\multirow{5}{*}{\texttt{OLMoE-7B}}
& FP & 16-16 & -- & 7.49 & 10.52 & 48.46 & 78.45 & 74.56 & 80.69 & 69.22 & 70.28 \\
\cline{2-12}

& DuQuant & 4-4 & C4 & 8.65 & 11.48 & 46.75 & 75.00 & 70.89 & 77.64 & 63.69 & 66.79 \\
& EAQuant & 4-4 & C4 &
\textbf{8.57} & \textbf{11.43} &
\textbf{47.61} & \textbf{75.92} & \textbf{72.51} &
\textbf{77.86} & \textbf{66.93} &
\textbf{68.17}~(\textcolor{red}{\textbf{+1.38}}) \\
\cdashline{2-12}

& DuQuant & 3-4 & C4 & 10.91 & 13.53 & 43.69 & 72.98 & 69.11 & 75.30 & 61.01 & 64.42 \\
& EAQuant & 3-4 & C4 &
\textbf{10.52} & \textbf{13.16} &
\textbf{44.71} & \textbf{73.99} & 67.86 &
\textbf{76.71} & \textbf{64.01} &
\textbf{65.46}~(\textcolor{red}{\textbf{+1.04}}) \\

\bottomrule
\end{tabular}
\vspace{-1em} \end{table*}


\subsection{Per-Expert Quantization Error Analysis}

Finally, Table~\ref{tab:layer15_per_expert_full} provides a complete per-expert quantization error (QE) comparison for Layer~15 of OLMoE-7B, with and without EA-CDB. The results show that EA-CDB reduces quantization error for the vast majority of experts, with especially large reductions observed for rarely activated experts (e.g., Experts 19, 32, and 33). While a small number of experts exhibit negligible increases in QE, the overall trend strongly favors EA-CDB, confirming that balanced calibration significantly improves quantization robustness at the expert level.

\begin{table}[!h]
\centering
\caption{Complete table of per-expert quantization error (QE) comparison for OLMoE Layer~15.
$\Delta$QE denotes the change in quantization error after applying EA-CDB,
computed as $\Delta$QE = QE$_{\text{w/ EA-CDB}}$ $-$ QE$_{\text{w/o EA-CDB}}$.
Negative values indicate reduced quantization error.
Lower QE corresponds to better quantization robustness.}
\label{tab:layer15_per_expert_full}
\small
\setlength{\tabcolsep}{4pt}\renewcommand{\arraystretch}{1.1}
\begin{tabular}{ccccc|ccccc}
\toprule
\multicolumn{5}{c|}{\textbf{Experts 0-31}} & \multicolumn{5}{c}{\textbf{Experts 32-63}} \\
\midrule
\textbf{ID} & \textbf{QE~(w/o EA-CDB)} & \textbf{QE~(w/ EA-CDB)} & \textbf{$\Delta$QE} & \textbf{Reduction} &
\textbf{ID} & \textbf{w/o EA-CDB} & \textbf{w/ EA-CDB} & \textbf{$\Delta$QE} & \textbf{Reduction} \\
\midrule
0  & 0.0195 & 0.0174 & -0.0021 & \textbf{10.8} &
32 & 0.0132 & 0.0060 & -0.0072 & \textbf{54.5} \\
1  & 0.0381 & 0.0303 & -0.0078 & \textbf{20.5} &
33 & 0.0092 & 0.0032 & -0.0060 & \textbf{65.2} \\
2  & 0.0283 & 0.0151 & -0.0132 & \textbf{46.6} &
34 & 0.0125 & 0.0077 & -0.0048 & \textbf{38.4} \\
3  & 0.0128 & 0.0101 & -0.0027 & \textbf{21.1} &
35 & 0.0107 & 0.0085 & -0.0022 & \textbf{20.6} \\
4  & 0.0179 & 0.0168 & -0.0011 & \textbf{6.1}  &
36 & 0.0257 & 0.0184 & -0.0073 & \textbf{28.4} \\
5  & 0.0104 & 0.0090 & -0.0014 & \textbf{13.5} &
37 & 0.0106 & 0.0064 & -0.0042 & \textbf{39.6} \\
6  & 0.0078 & 0.0070 & -0.0008 & \textbf{10.3} &
38 & 0.0110 & 0.0098 & -0.0012 & \textbf{10.9} \\
7  & 0.0372 & 0.0361 & -0.0011 & \textbf{3.0}  &
39 & 0.0105 & 0.0086 & -0.0019 & \textbf{18.1} \\
8  & 0.0075 & 0.0059 & -0.0016 & \textbf{21.3} &
40 & 0.0134 & 0.0106 & -0.0028 & \textbf{20.9} \\
9  & 0.0086 & 0.0058 & -0.0028 & \textbf{32.6} &
41 & 0.0173 & 0.0162 & -0.0011 & \textbf{6.4}  \\
10 & 0.0179 & 0.0155 & -0.0024 & \textbf{13.4} &
42 & 0.0101 & 0.0086 & -0.0015 & \textbf{14.9} \\
11 & 0.0104 & 0.0076 & -0.0028 & \textbf{26.9} &
43 & 0.0109 & 0.0093 & -0.0016 & \textbf{14.7} \\
12 & 0.0151 & 0.0123 & -0.0028 & \textbf{18.5} &
44 & 0.0091 & 0.0056 & -0.0035 & \textbf{38.5} \\
13 & 0.0270 & 0.0187 & -0.0083 & \textbf{30.7} &
45 & 0.0111 & 0.0102 & -0.0009 & \textbf{8.1}  \\
14 & 0.0116 & 0.0108 & -0.0008 & \textbf{6.9}  &
46 & 0.0101 & 0.0094 & -0.0007 & \textbf{6.9}  \\
15 & 0.0147 & 0.0142 & -0.0005 & \textbf{3.4}  &
47 & 0.0209 & 0.0175 & -0.0034 & \textbf{16.3} \\
16 & 0.0096 & 0.0073 & -0.0023 & \textbf{24.0} &
48 & 0.0083 & 0.0071 & -0.0012 & \textbf{14.5} \\
17 & 0.0161 & 0.0108 & -0.0053 & \textbf{32.9} &
49 & 0.0145 & 0.0139 & -0.0006 & \textbf{4.1}  \\
18 & 0.0160 & 0.0101 & -0.0059 & \textbf{36.9} &
50 & 0.0095 & 0.0074 & -0.0021 & \textbf{22.1} \\
19 & 0.0134 & 0.0057 & -0.0077 & \textbf{57.5} &
51 & 0.0095 & 0.0080 & -0.0015 & \textbf{15.8} \\
20 & 0.0089 & 0.0080 & -0.0009 & \textbf{10.1} &
52 & 0.0258 & 0.0232 & -0.0026 & \textbf{10.1} \\
21 & 0.0152 & 0.0134 & -0.0018 & \textbf{11.8} &
53 & 0.0076 & 0.0066 & -0.0010 & \textbf{13.2} \\
22 & 0.0074 & 0.0068 & -0.0006 & \textbf{8.1}  &
54 & 0.0164 & 0.0128 & -0.0036 & \textbf{22.0} \\
23 & 0.0108 & 0.0102 & -0.0006 & \textbf{5.6}  &
55 & 0.0102 & 0.0103 & +0.0001 & -1.0 \\
24 & 0.0111 & 0.0083 & -0.0028 & \textbf{25.2} &
56 & 0.0205 & 0.0166 & -0.0039 & \textbf{19.0} \\
25 & 0.0129 & 0.0121 & -0.0008 & \textbf{6.2}  &
57 & 0.0069 & 0.0064 & -0.0005 & \textbf{7.2}  \\
26 & 0.0142 & 0.0138 & -0.0004 & \textbf{2.8}  &
58 & 0.0204 & 0.0200 & -0.0004 & \textbf{2.0}  \\
27 & 0.0095 & 0.0080 & -0.0015 & \textbf{15.8} &
59 & 0.0120 & 0.0113 & -0.0007 & \textbf{5.8}  \\
28 & 0.0223 & 0.0205 & -0.0018 & \textbf{8.1}  &
60 & 0.0262 & 0.0220 & -0.0042 & \textbf{16.0} \\
29 & 0.0202 & 0.0205 & +0.0003 & -1.5 &
61 & 0.0075 & 0.0073 & -0.0002 & \textbf{2.7}  \\
30 & 0.0098 & 0.0094 & -0.0004 & \textbf{4.1}  &
62 & 0.0109 & 0.0113 & +0.0004 & -3.7 \\
31 & 0.0116 & 0.0108 & -0.0008 & \textbf{6.9}  &
63 & 0.0098 & 0.0083 & -0.0015 & \textbf{15.3} \\
\bottomrule
\end{tabular}
\vspace{-1em}
\end{table}

\section{Additional Visual Evidence for Expert-Aware Design}
\label{sec:appx_vis}

This appendix provides extended visual analyses corresponding to Figures~\ref{fig_observation}(a), \ref{fig_observation}(b), and \ref{fig_observation}(c) in the main paper.
These results offer additional empirical evidence supporting the design of the three core components of EAQuant: Expert-Aware Smoothing Aggregation (EA-SA), Expert-Aware Routing Consistency Alignment (EA-RCA), and Expert-Aware Calibration Data Balance (EA-CDB).

\subsection{Activation Distribution Visualization}

\begin{figure*}[!h]
	\centering
	\includegraphics[width=\linewidth]{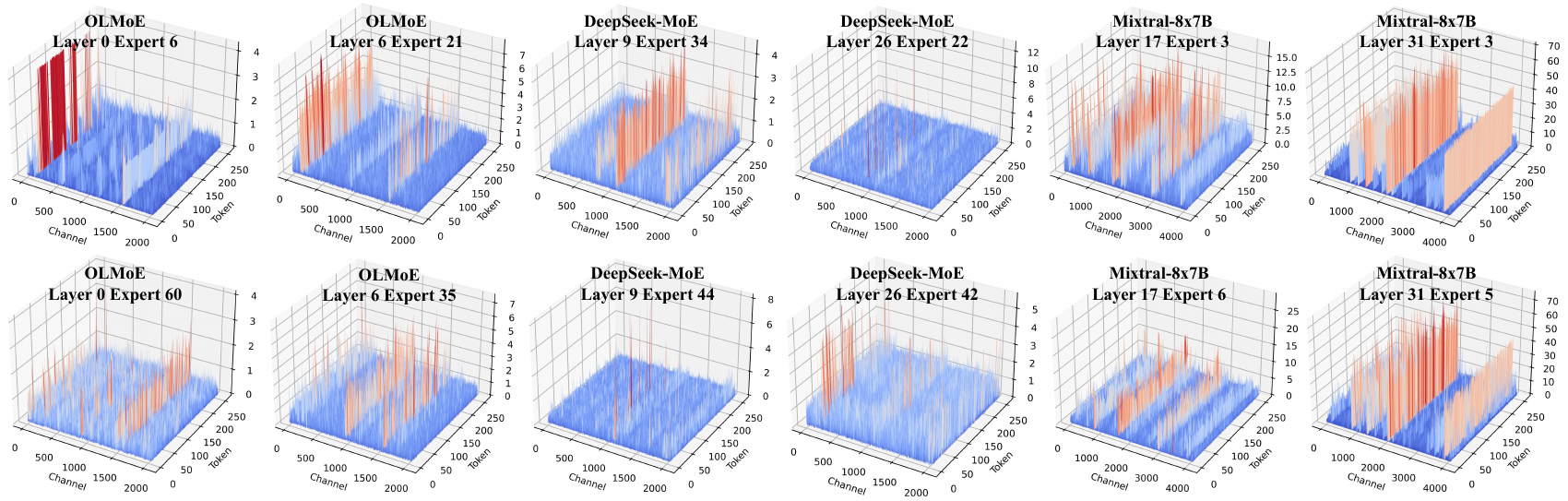}
	\vspace{-2em}
	\caption{Extended visualization for Figure~\ref{fig_Activation_distribution}: Activation outliers across channels in different experts and layers of three MoE models, exhibiting similar distribution patterns to dense models and posing challenges for quantization.}
	\label{fig_Activation_distribution_all}
\end{figure*}

Figure~\ref{fig_Activation_distribution_all} provides an extended visualization of activation distributions across channels for different experts and layers in three representative MoE models: OLMoE-7B, DeepSeek-MoE-16B, and Mixtral-8x7B. This figure serves as a comprehensive version of Figure~\ref{fig_Activation_distribution} presented in the main paper, where we focus on OLMoE-7B for clarity and space considerations. These observations suggest that activation outliers are a ubiquitous phenomenon in MoE models and persist across scales and architectures, posing significant challenges for low-bit post-training quantization. This motivates the need for expert-aware activation smoothing strategies, such as those proposed in EAQuant, to robustly handle channel-wise extremes across all experts and layers.

\subsection{Channel-wise Similarity across Experts and EA-SA}

\begin{figure*}[!h]
	\centering
	\includegraphics[width=\linewidth]{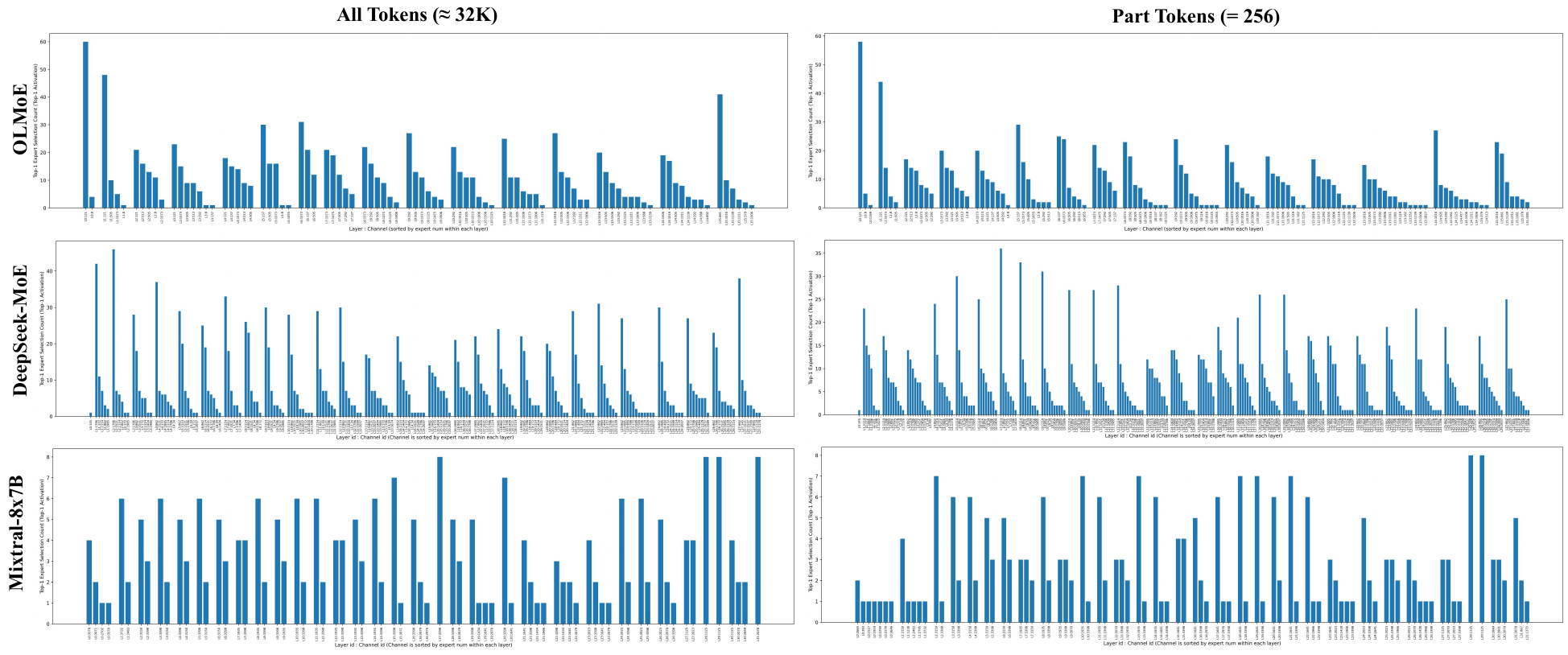}
	\caption{Extended visualization for Figure~\ref{fig_observation}(a): channel-wise concentration of dominant activations across \emph{all} experts in MoE models.}
	\label{fig:supp_ea_sa}
\end{figure*}

Figure~\ref{fig:supp_ea_sa} provides an extended analysis of Figure~\ref{fig_observation}(a) by visualizing channel-wise dominant activations for \emph{all experts} in three representative MoE models, including OLMoE-7B, DeepSeek-MoE-16B, and Mixtral-8x7B. 
While Figure~\ref{fig_observation}(a) in the main paper illustrates this phenomenon using a subset of experts for clarity, the supplementary results confirm that the observed patterns consistently hold across the full set of experts and corresponding MoE layers in each model.

Despite sparse and input-dependent expert activation, we observe strong channel-wise concentration and substantial overlap in dominant activation dimensions across different experts.
This indicates that activation outliers are not randomly distributed or uniquely tied to individual experts, but are instead aligned along a shared subset of channel dimensions.
Such cross-expert channel-wise similarity provides strong empirical justification for Expert-Aware Smoothing Aggregation (EA-SA), which leverages a unified channel-wise smoothing vector shared across experts to effectively suppress activation outliers while preserving exact functional equivalence.

\subsection{Routing Sensitivity under Quantization and EA-RCA}

\begin{figure*}[!h]
	\centering
	\includegraphics[width=\linewidth]{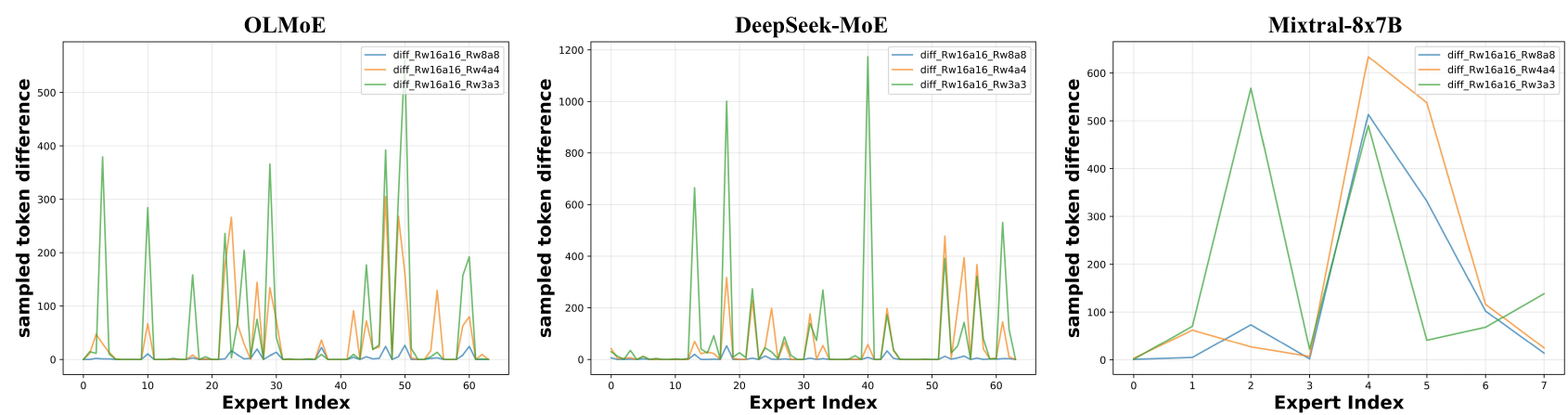}
	\caption{Extended visualization for Figure~\ref{fig_observation}(b): quantization sensitivity of router logits and expert selection in MoE models.}
	\label{fig:supp_ea_rca}
\end{figure*}

Figure~\ref{fig:supp_ea_rca} extends Figure~\ref{fig_observation}(b) by illustrating the quantization sensitivity of router logits and expert selection across three MoE models.
The results show that small quantization-induced perturbations in router outputs can significantly alter top-$k$ expert assignment, leading to unstable routing behavior.
This phenomenon is consistently observed across different architectures, motivating Expert-Aware Routing Consistency Alignment (EA-RCA), which explicitly preserves routing behavior during post-training quantization.

\subsection{Expert Activation Imbalance and EA-CDB}

\begin{figure*}[!h]
	\centering
	\includegraphics[width=\linewidth]{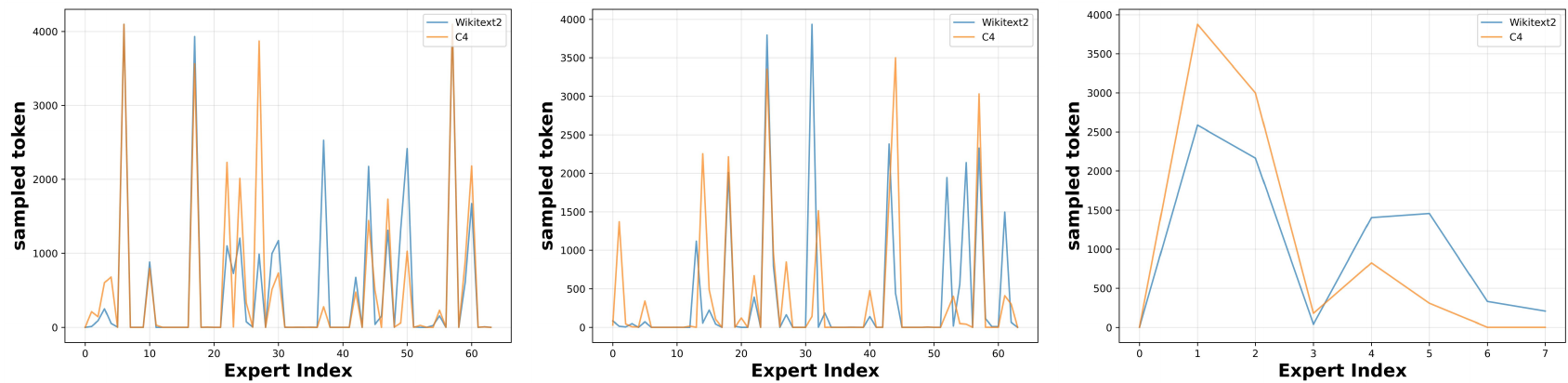}
	\caption{Extended visualization for Figure~\ref{fig_observation}(c): expert activation imbalance and calibration data distribution in MoE models.}
	\label{fig:supp_ea_cdb}
\end{figure*}

Figure~\ref{fig:supp_ea_cdb} extends Figure~\ref{fig_observation}(c) by presenting expert activation statistics for OLMoE-7B, DeepSeek-MoE-16B, and Mixtral-8x7B.
Across all models, expert activation follows a highly skewed long-tailed distribution, where a small subset of experts receives the majority of tokens while many experts are sparsely activated.
Such imbalance leads to insufficient calibration data coverage for under-utilized experts and motivates the proposed Expert-Aware Calibration Data Balance (EA-CDB), which explicitly rebalances calibration samples to improve expert-level quantization robustness.

\end{document}